\documentclass[11pt]{article}
\usepackage[margin=1in]{geometry}
\usepackage{amsmath,amssymb,amsfonts}
\usepackage{algorithm}
\usepackage{algorithmic}
\usepackage{graphicx}
\usepackage{xcolor}
\usepackage{tikz}
\usetikzlibrary{arrows.meta,positioning,calc,shapes.geometric,fit,backgrounds}
\usepackage{float}
\usepackage{placeins}
\usepackage{array}
\usepackage{booktabs}
\usepackage{multirow}
\usepackage{tabularx}
\usepackage{ragged2e}
\usepackage{graphicx}
\usepackage[font=small,labelfont=bf]{caption}
\usepackage{url}
\usepackage[hidelinks]{hyperref}
\newcolumntype{L}[1]{>{\RaggedRight\arraybackslash}p{#1}}
\newcolumntype{Y}{>{\RaggedRight\arraybackslash}X}

\tikzset{
    procbox/.style={
        rectangle, rounded corners, draw=black, thick,
        fill=blue!5,
        align=center,
        inner sep=5pt, minimum height=9mm, minimum width=22mm,
        font=\small
    },
    statebox/.style={
        rectangle, draw=black, thick,
        fill=gray!8,
        align=center,
        inner sep=5pt, minimum height=8mm,
        font=\small
    },
    toolbox/.style={
        rectangle, rounded corners, draw=black,
        fill=green!6,
        align=center,
        inner sep=4pt, minimum height=7mm,
        font=\small
    },
    frontierbox/.style={
        rectangle, rounded corners, draw=black, very thick,
        fill=orange!18,
        align=center,
        inner sep=4pt, minimum height=7mm,
        font=\small
    },
    flow/.style={-{Stealth[length=2.4mm]}, thick},
    edgelbl/.style={font=\footnotesize, midway}
}
\title{ToolChoiceConfusion: Causal Minimal Tool Filtering for Reliable LLM Agents}
\author{
Rahul Suresh Babu\\
Independent Researcher\\
United States of America\\
\texttt{rahulsb@bu.edu}
\and
Laxmipriya Ganesh Iyer\\
Independent Researcher\\
United States of America\\
\texttt{iyer.la@northeastern.edu}
}
\date{}
\begin{document}
\maketitle
\begin{abstract}
Large language model agents increasingly rely on external tools, but larger tool menus can reduce reliability and efficiency by increasing wrong-tool calls, premature actions, and token cost. Existing tool-selection methods often optimize semantic relevance, exposing tools whose names or descriptions match the user request. We argue that relevance is insufficient: a tool may be related to the task while still being unnecessary or premature at the current step.

We propose \textit{Causal Minimal Tool Filtering} (CMTF), a training-free method that selects tools by causal sufficiency. CMTF uses lightweight precondition-effect contracts to expose only the minimal next-step tool frontier needed to advance from the current state toward the user goal. Across multi-step tool-use tasks, we compare CMTF with all-tools exposure, keyword retrieval, state-aware filtering, and causal-path ablations, measuring task success, wrong-tool calls, premature actions, tool exposure, and token cost. In the main benchmark with 102 tasks, 100 tools, four LLM backends, and 2448 task-method-model runs, CMTF matches the strongest causal baseline in aggregate success while reducing visible tools from 100 to one per step and reducing token usage by about 90\% relative to all-tools exposure.
\end{abstract}

\section{Introduction}
Tool access has become a central mechanism for extending large language models (LLMs) beyond text generation. Modern LLM agents can interleave reasoning with external actions, invoke APIs, search information sources, operate over files, update calendars, draft emails, execute code, and interact with structured systems~\cite{yao2022react,schick2023toolformer}. This shift has motivated a growing body of work on tool-use training, function-calling evaluation, and tool-use benchmarks~\cite{qin2023toollm,patil2024bfcl}. As tool ecosystems grow, however, an agent must solve a second problem before it can use a tool correctly: it must decide which tools should be visible at each step of the task.
A common approach is to treat tool selection as a relevance problem. Given a user request, retrieval or filtering methods expose tools whose names, descriptions, or schemas appear most related to the query~\cite{shi2025toolret,gan2025ragmcp}. Recent work has also studied the effect of shortlist size, showing that exposing too many tools can increase selection difficulty while exposing too few may omit the correct tool~\cite{repantis2026howmanytools}. Other systems reduce ambiguity by merging redundant tools or filtering context-aware candidates~\cite{liu2025toolscope}. These approaches address an important scalability challenge, but they largely preserve a relevance-oriented view of tool exposure.
In this paper, we argue that semantic relevance is not sufficient for reliable multi-step tool use. A tool may be related to the user request while still being unnecessary, premature, or distracting at the current step. For example, in a task such as ``find an email and draft a reply,'' tools for searching email, reading email, creating a draft, sending an email, and archiving a message are all semantically related. Yet only the search tool is causally useful before a message identifier is known, and exposing later-stage or high-risk tools too early can lead to wrong-tool calls, premature actions, longer trajectories, and higher token cost. We refer to this failure mode as \textit{ToolChoiceConfusion}: degradation in agent behavior caused by exposing tools that are plausible but not causally necessary for the current decision.
We propose \textit{Causal Minimal Tool Filtering} (CMTF), a training-free method for selecting tools based on causal sufficiency rather than semantic relevance alone. CMTF represents each tool using a lightweight contract consisting of preconditions and effects. Given the current task state and the user goal, it constructs a dependency graph over possible state transitions and exposes only the minimal next-step tool frontier needed to advance toward the goal. We use the term causal in an operational sense: a tool is causally relevant when its effects help transform the current state into a state required for task completion.
This framing distinguishes CMTF from simpler filtering strategies. All-tools exposure maximizes availability but increases selection burden. Keyword or embedding retrieval may return related tools without regard to when they should be used. State-aware filtering removes tools whose inputs are unavailable, but it can still expose tools that are executable yet not goal-directed. CMTF instead asks whether a tool lies on a minimal causal path from the current state to the goal, and exposes only the next frontier of that path.
We evaluate this idea on controlled multi-step tool-use tasks with synthetic tool registries and mocked tool outputs. The benchmark is designed to isolate tool-choice behavior rather than external API variability. We compare CMTF against all-tools exposure, keyword-based top-$k$ retrieval, state-aware filtering, and causal-path ablations that expose future-needed tools upfront. We measure final task success, wrong-tool calls, premature actions, tool exposure, trajectory length, and token cost.
This paper makes the following contributions:
\begin{enumerate}
    \item We formulate \textit{ToolChoiceConfusion} as a tool-exposure problem in multi-step LLM agents, where semantically plausible tools can degrade reliability when they are not causally necessary at the current step.
    \item We introduce \textit{Causal Minimal Tool Filtering} (CMTF), a lightweight, training-free filtering method based on tool preconditions, effects, task state, and goal state.
    \item We define evaluation metrics for causal tool filtering, including wrong-tool rate, premature action rate, tool exposure, trajectory length, token cost, and causal frontier behavior.
    \item We construct a controlled synthetic benchmark for evaluating tool filtering in multi-step agent tasks across multiple domains.
    \item We empirically compare CMTF with all-tools exposure, keyword retrieval, state-aware filtering, and causal-path ablations to test whether agents benefit from seeing fewer, causally sufficient tools at each step.
\end{enumerate}
\section{Background and Related Work}
\label{sec:background}

\subsection{Tool-Augmented LLM Agents}

Tool use has become a central mechanism for extending large language models beyond text generation. ReAct introduced interleaved reasoning and acting~\cite{yao2022react}, Toolformer showed that models can learn to invoke external APIs~\cite{schick2023toolformer}, and ToolLLM/ToolBench scaled tool-use evaluation to large API ecosystems~\cite{qin2023toollm}. These systems establish tool use as a core capability for LLM agents, but they also create a systems problem: as tool libraries grow, the agent must decide which tools should be visible at each step.

\subsection{Function Calling and Tool-Use Evaluation}

Benchmarks such as API-Bank~\cite{li2023apibank}, the Berkeley Function-Calling Leaderboard~\cite{patil2024bfcl}, and AgentBench~\cite{liu2023agentbench} evaluate tool-use capabilities including API selection, argument construction, multi-turn use, and interactive task completion. These benchmarks are important for measuring whether models can call tools correctly, but they typically assume that the available tool interface has already been defined. In contrast, this paper focuses on the upstream tool-exposure problem: which tools should be visible before each decision?

\subsection{Reliability and Runtime Orchestration}

Recent work also studies reliability in tool-augmented agents as a runtime orchestration problem. Babu and Agrawal~\cite{babu2026selfhealing} propose a self-healing orchestrator that monitors execution, detects failures, selects recovery actions, and verifies recovered trajectories. Our work is complementary: rather than recovering from tool-use failures after execution, CMTF reduces the likelihood of tool-choice errors before execution by controlling the visible tool set at each decision step.

\subsection{Tool Retrieval, Pruning, and Ambiguity}

As tool libraries expand, retrieval and pruning become necessary for both efficiency and reliability. Recent work studies tool retrieval at scale~\cite{shi2025toolret}, retrieval-augmented tool selection to reduce prompt bloat~\cite{gan2025ragmcp}, shortlist-size tradeoffs~\cite{repantis2026howmanytools}, and ambiguity from redundant or overlapping tools~\cite{liu2025toolscope}. These approaches show that tool menus are not neutral context: the visible tool set can affect cost, selection difficulty, and downstream behavior.

However, most retrieval-based methods still treat filtering as a relevance or shortlist-selection problem. They ask which tools are semantically related to the user request or how many candidates should be shown. CMTF instead asks whether a tool is causally needed at the current step. A tool can be relevant to the overall task while still being premature or distracting before its preconditions are satisfied.

\subsection{Preconditions, Effects, and Causal Tool Exposure}

CMTF is inspired by the precondition-effect abstraction used in classical planning. STRIPS represents actions in terms of the conditions required before execution and the effects produced afterward, enabling planners to search for action sequences that transform an initial state into a goal state~\cite{fikes1971strips}. PDDL later standardized planning-domain representations around related notions of states, actions, and goals~\cite{mcdermott1998pddl}. We use this abstraction in a lightweight way: each tool is represented by the state variables it requires and the state variables it produces.

Unlike a full symbolic planner, CMTF does not replace the LLM agent. It filters the tool menu exposed to the agent. This framing separates CMTF from three common alternatives. Relevance-based retrieval may expose tools that sound related but are not needed now. State-aware filtering may expose tools that are executable but not goal-directed. Full causal-path exposure may reveal future-needed tools too early. CMTF instead exposes only the minimal next-step frontier on a precondition-effect dependency graph, making tool filtering a question of causal sufficiency rather than semantic relevance alone.

\section{Problem Formulation}
\label{sec:problem}
We study tool selection in multi-step LLM agents. At each step, an agent receives a user task, a current task state, and a visible subset of tools. The agent selects one tool, observes its output, and updates its state. The central question is how to construct the visible tool set so that the agent can make progress without being distracted by irrelevant, premature, or non-goal-directed tools.
\subsection{Tools, State, and Goals}
Let $\mathcal{T} = \{t_1, t_2, \ldots, t_n\}$ denote the full tool library. Each tool $t_i$ is represented as a lightweight contract:
\[
t_i = (d_i, R_i, E_i, c_i, \rho_i),
\]
where $d_i$ is a natural-language description, $R_i$ is the set of required state variables, $E_i$ is the set of state variables produced by the tool, $c_i$ is an optional cost, and $\rho_i$ is an optional risk level. The description $d_i$ supports relevance-based filtering, while the precondition-effect fields $(R_i, E_i)$ support causal filtering.
Let $\mathcal{X}$ be the universe of possible state variables. At step $t$, the task state is a set of known variables $s_t \subseteq \mathcal{X}$. In implementation, state variables may also have concrete values used for tool arguments, but the filtering logic reasons over variable availability. A user task is associated with a goal state $g \subseteq \mathcal{X}$, and the task is complete when $g \subseteq s_t$.
At each step, a filtering method selects a visible tool set
\[
\mathcal{V}_t \subseteq \mathcal{T}.
\]
The agent chooses a tool $a_t \in \mathcal{V}_t$, receives an observation, and the symbolic state is updated as:
\[
s_{t+1} = s_t \cup E_{a_t}.
\]
The process repeats until the goal is reached, a failure occurs, or a maximum step limit is exceeded.
\subsection{Relevance, Executability, and Causal Sufficiency}
We distinguish three notions of tool suitability. A tool is \textit{relevant} if its name or description is semantically related to the user request. A tool is \textit{executable} at state $s_t$ if its required variables are available:
\[
R_i \subseteq s_t.
\]
A tool is \textit{causally sufficient} at state $s_t$ if it is executable and lies on a valid dependency path from the current state to the goal. Thus, a tool may be relevant but not executable, executable but not goal-directed, or useful later but premature at the current step.
A causal path is a sequence of tools
\[
\pi = (t_1, t_2, \ldots, t_k)
\]
such that each tool's requirements are satisfied by the initial state and the effects of earlier tools:
\[
R_{t_j} \subseteq s_t \cup \bigcup_{\ell < j} E_{t_\ell},
\]
and the final accumulated state satisfies the goal:
\[
g \subseteq s_t \cup \bigcup_{j=1}^{k} E_{t_j}.
\]
A minimal causal path is a valid path with minimum length or cost. The \textit{next causal frontier} is the first executable tool or set of tools on such a path.
\subsection{ToolChoiceConfusion and Objective}
We define \textit{ToolChoiceConfusion} as degraded agent behavior caused by exposing tools that are plausible but not appropriate for the current decision. This includes irrelevant tools, tools that are useful only at future steps, executable tools that do not advance the goal, and high-risk tools exposed without causal need. These exposures can lead to wrong-tool calls, premature actions, longer trajectories, higher token cost, and lower task success.
The objective is to select, at each step, a visible tool set $\mathcal{V}_t$ that preserves progress toward the goal while minimizing unnecessary exposure. Ideally, $\mathcal{V}_t$ should contain the next causally sufficient frontier and exclude tools that are irrelevant, premature, redundant, or executable but not goal-directed. CMTF approximates this objective by constructing a precondition-effect dependency graph, finding a minimal causal path from $s_t$ to $g$, and exposing only the next executable frontier of that path.
\section{Causal Minimal Tool Filtering}
\label{sec:cmtf}
Causal Minimal Tool Filtering (CMTF) is a lightweight tool-exposure method for multi-step LLM agents. Given a current task state $s_t$, a goal state $g$, and a tool library $\mathcal{T}$, CMTF uses tool preconditions and effects to identify the minimal next-step tool frontier needed to advance toward the goal. The LLM agent still chooses tool arguments and executes the task step by step; CMTF only controls which tools are visible for each local decision.
\begin{figure}[t]
    \centering
    \begin{tikzpicture}[node distance=7mm and 9mm]
        \node[procbox] (inputs) {Current state $s_t$ \\ + goal $g$ \\ + tool contracts};
        \node[procbox, right=of inputs] (graph) {Dependency \\ graph};
        \node[procbox, right=of graph] (path) {Minimal \\ causal path};
        \node[procbox, below=of path] (frontier) {Next causal \\ frontier};
        \node[procbox, left=of frontier] (menu) {Visible \\ tool menu};
        \node[procbox, left=of menu] (call) {Tool call \\ + state update};
        \draw[flow] (inputs) -- (graph);
        \draw[flow] (graph) -- (path);
        \draw[flow] (path) -- (frontier);
        \draw[flow] (frontier) -- (menu);
        \draw[flow] (menu) -- (call);
        \draw[flow, dashed, blue!55!black]
            (call.west) -- ($(call.west)+(-7mm,0)$)
            |- node[edgelbl, fill=white, inner sep=2pt, pos=0.25, sloped]
                {feedback: updated state $s_{t+1}$}
            (inputs.west);
    \end{tikzpicture}
    \caption{Overview of CMTF. The filter constructs a precondition-effect dependency graph, identifies a minimal causal path from the current state to the goal, and exposes only the next executable frontier to the LLM agent. The updated state is fed back for the next step.}
    \label{fig:cmtf_pipeline}
\end{figure}
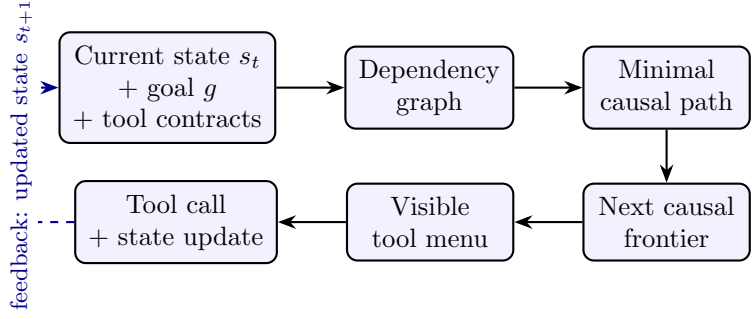
\subsection{Dependency Graph and Path Selection}
Each tool $t_i$ has required variables $R_i$ and produced variables $E_i$. CMTF treats an applicable tool as a state transition:
\[
s \xrightarrow{t_i} s \cup E_i
\quad \text{if} \quad R_i \subseteq s.
\]
The resulting dependency graph defines which tools can produce the state variables needed to reach the goal. A causal path is a sequence of tools
\[
\pi = (t_1, t_2, \ldots, t_k)
\]
such that the accumulated effects of the sequence satisfy the goal:
\[
g \subseteq s_t \cup \bigcup_{j=1}^{k} E_{t_j}.
\]
CMTF selects a minimal valid path. In the main experiments, minimality is defined by path length, corresponding to unit tool costs. More generally, path selection can incorporate cost or risk penalties:
\[
\operatorname{score}(\pi) =
\sum_{t_i \in \pi} c_i
+
\lambda \sum_{t_i \in \pi} \operatorname{risk}(\rho_i).
\]
\subsection{Frontier-Based Tool Exposure}
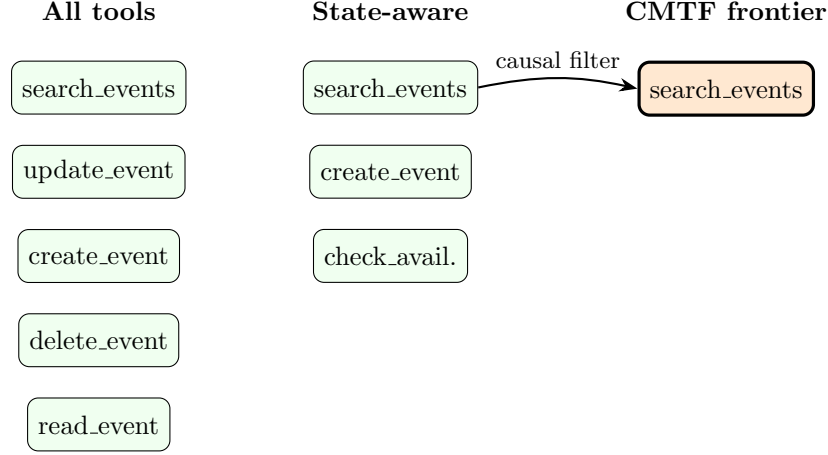
\begin{figure}[t]
    \centering
    \begin{tikzpicture}[node distance=4mm and 6mm]
        \node[font=\small\bfseries] (h1) {All tools};
        \node[font=\small\bfseries, right=18mm of h1] (h2) {State-aware};
        \node[font=\small\bfseries, right=18mm of h2] (h3) {CMTF frontier};
        \node[toolbox, below=of h1] (a1) {search\_events};
        \node[toolbox, below=of a1] (a2) {update\_event};
        \node[toolbox, below=of a2] (a3) {create\_event};
        \node[toolbox, below=of a3] (a4) {delete\_event};
        \node[toolbox, below=of a4] (a5) {read\_event};
        \node[toolbox, below=of h2] (b1) {search\_events};
        \node[toolbox, below=of b1] (b2) {create\_event};
        \node[toolbox, below=of b2] (b3) {check\_avail.};
        \node[frontierbox, below=of h3] (c1) {search\_events};
        \draw[flow] (b1.east) to[bend left=12]
            node[edgelbl, above, pos=0.5] {causal filter}
            (c1.west);
    \end{tikzpicture}
    \caption{Tool exposure for the first step of ``move tomorrow's dentist appointment.'' All-tools exposure shows the whole calendar registry; state-aware filtering keeps every executable tool; CMTF exposes only \texttt{search\_events}, the single tool on the minimal causal path. Box labels use a uniform size; the edge annotation uses the smaller arrow-label size.}
    \label{fig:exposure_comparison}
\end{figure}

CMTF exposes only the next executable frontier of the selected path. If the minimal path is
\[
\pi^\star = (t_1^\star, t_2^\star, \ldots, t_k^\star),
\]
then the visible tool set is:
\[
\mathcal{V}_t = \{t_1^\star\}.
\]
If multiple minimal paths are tied, the frontier may contain multiple first-step tools. In our controlled experiments, we use deterministic tie-breaking and expose one frontier tool.
This differs from common alternatives. All-tools exposure shows the entire registry. Relevance-based retrieval shows tools that match the request text. State-aware filtering shows all executable tools, even if they do not advance the goal. Full-causal-path exposure shows all tools on a valid path, including future-step tools that may be premature. CMTF instead exposes only the tool needed for the current causal step.
Figure~\ref{fig:exposure_comparison} contrasts these exposure strategies on the running calendar example.
\subsection{Algorithm}
Algorithm~\ref{alg:cmtf} summarizes the breadth-first version of CMTF. The algorithm searches over accumulated symbolic states and returns the first tool on a shortest path to the goal. If no path is found, the controlled benchmark treats this as a filter failure; in deployment, this case could fall back to clarification or retrieval-based filtering.
\begin{algorithm}[t]
\caption{Causal Minimal Tool Filtering}
\label{alg:cmtf}
\begin{algorithmic}[1]
\REQUIRE Current state $s_t$, goal state $g$, tool library $\mathcal{T}$
\ENSURE Visible tool set $\mathcal{V}_t$
\IF{$g \subseteq s_t$}
    \STATE \textbf{return} $\emptyset$
\ENDIF
\STATE Initialize queue $Q \leftarrow [(s_t, [\ ])]$
\STATE Initialize visited set $U \leftarrow \{s_t\}$
\WHILE{$Q$ is not empty}
    \STATE Pop $(s, \pi)$ from $Q$
    \IF{$g \subseteq s$}
        \STATE $\pi^\star \leftarrow \pi$
        \STATE \textbf{return} first executable frontier of $\pi^\star$
    \ENDIF
    \FORALL{tools $t_i \in \mathcal{T}$}
        \IF{$R_i \subseteq s$}
            \STATE $s' \leftarrow s \cup E_i$
            \IF{$s' \notin U$}
                \STATE Add $(s', \pi \mathbin{\Vert} [t_i])$ to $Q$
                \STATE Add $s'$ to $U$
            \ENDIF
        \ENDIF
    \ENDFOR
\ENDWHILE
\STATE \textbf{return} $\emptyset$
\end{algorithmic}
\end{algorithm}
\subsection{Running Example}
Consider the task: \textit{``Move tomorrow's dentist appointment to 4 PM.''} The initial state is:
\[
s_0 = \{\texttt{date}, \texttt{event\_description}, \texttt{new\_time}\},
\]
and the goal is:
\[
g = \{\texttt{event\_updated}\}.
\]
A calendar tool registry may include \texttt{search\_events}, \texttt{read\_event}, \texttt{update\_event}, \texttt{create\_event}, and \texttt{delete\_event}. Although these tools are all semantically related to the task, only \texttt{search\_events} produces the missing \texttt{event\_id} needed to update the existing event. CMTF therefore identifies the path:
\[
\texttt{search\_events} \rightarrow \texttt{update\_event}.
\]
At the first step, CMTF exposes only:
\[
\mathcal{V}_0 = \{\texttt{search\_events}\}.
\]
After \texttt{event\_id} is produced, CMTF exposes:
\[
\mathcal{V}_1 = \{\texttt{update\_event}\}.
\]
This example illustrates the central distinction: tools such as \texttt{create\_event} and \texttt{delete\_event} are relevant to calendar operations, but they are not on the minimal causal path for the requested task.
\FloatBarrier
\section{Benchmark Design}
\label{sec:benchmark}

We construct a controlled synthetic benchmark to isolate the effect of tool-menu construction on multi-step LLM agent behavior. The benchmark is diagnostic rather than a full simulation of real-world agents: all tool outputs are mocked and deterministic, allowing us to attribute failures to tool selection rather than external API variability. The benchmark is designed to test whether a filtering method can expose the correct next tool while avoiding four common confusers: semantically related but wrong tools, future-step tools exposed too early, executable but non-goal-directed tools, and high-risk actions such as send, update, share, or delete.

\subsection{Domains and Task Patterns}

The benchmark focuses on three workflow domains: calendar, email, and files/documents. These domains naturally induce multi-step trajectories such as search-read-write and search-read-summarize, while also containing realistic high-risk distractors. The main benchmark contains 102 tasks, with 34 tasks per domain. Tasks are generated by varying the requested operation, entity description, and task-specific state variables while preserving controlled gold tool chains. Each task specifies a natural-language query, initial state, goal state, gold tool chain, mocked tool outputs, and success criteria.

\begin{table}[t]
\centering
\small
\begin{tabularx}{\linewidth}{l L{3.2cm} Y}
\toprule
Domain & Task patterns & Example gold chain \\
\midrule
Calendar & move event, summarize event, invite attendee
& \texttt{search\_events} $\rightarrow$ \texttt{update\_event} \\
Email & draft reply, summarize email, extract deadline
& \texttt{search\_emails} $\rightarrow$ \texttt{read\_email} $\rightarrow$ \texttt{create\_draft} \\
Files & summarize section, summarize document, extract section
& \texttt{search\_files} $\rightarrow$ \texttt{read\_file} $\rightarrow$ \texttt{summarize\_section} \\
\bottomrule
\end{tabularx}
\caption{Benchmark domains and representative task patterns. The full benchmark contains multiple task instances per pattern, each with a gold tool chain and deterministic mocked tool outputs.}
\label{tab:benchmark_domains}
\end{table}

\subsection{Tool Registry and Distractors}

The tool registry contains 100 synthetic tools, consisting of core workflow tools and controlled distractors. Each tool includes a natural-language description, input schema, required state variables, produced state variables, risk level, and optional cost. The registry includes both task-relevant tools and distractors designed to stress tool-choice behavior:

\begin{itemize}
    \item \textbf{Relevant but wrong tools}: tools such as \texttt{create\_event} when the task requires updating an existing event.
    \item \textbf{Premature tools}: tools such as \texttt{update\_event} before an \texttt{event\_id} is known.
    \item \textbf{Near-duplicate tools}: tools such as \texttt{search\_emails} and \texttt{search\_email\_ids}.
    \item \textbf{Risky tools}: tools such as \texttt{send\_email}, \texttt{delete\_email}, \texttt{delete\_file}, and \texttt{share\_file}.
    \item \textbf{Cross-domain distractors}: tools from unrelated operational domains such as finance, web, support, analytics, payments, security, and database workflows.
\end{itemize}

This design stresses the central failure mode studied in this paper: a tool can be semantically plausible, executable, or useful in another context while still being inappropriate for the current decision.

\subsection{Mocked Execution}

Tool execution is simulated with deterministic outputs. When the agent selects a tool, the environment returns the predefined output for that task-tool pair if one exists; otherwise, it returns a controlled error. Produced variables are added to the task state, and the loop continues until the goal is reached, a failure occurs, or the maximum step limit is exceeded. Each run produces step-level traces containing the task identifier, model, filtering method, visible tools, selected tools, gold next tools, state transitions, token usage, and error status.

\subsection{Example Task}

Table~\ref{tab:benchmark_example} shows a representative calendar task. Several tools are calendar-related, but only \texttt{search\_events} is causally useful at the first step because it produces the missing \texttt{event\_id} required by \texttt{update\_event}.

\begin{table}[t]
\centering
\small
\begin{tabularx}{\linewidth}{l Y}
\toprule
Field & Value \\
\midrule
User query & Move tomorrow's dentist appointment to 4 PM \\
Initial state & \texttt{date}, \texttt{event\_description}, \texttt{new\_time} \\
Goal state & \texttt{event\_updated} \\
Gold chain & \texttt{search\_events} $\rightarrow$ \texttt{update\_event} \\
Relevant distractors & \texttt{create\_event}, \texttt{read\_event}, \texttt{delete\_event}, \texttt{check\_availability} \\
Mock output & \texttt{search\_events} produces \texttt{event\_id} \\
\bottomrule
\end{tabularx}
\caption{Example benchmark task. The distractor tools are calendar-related but not the minimal next causal step.}
\label{tab:benchmark_example}
\end{table}

\subsection{Scope}

The benchmark intentionally isolates tool-exposure behavior. It does not test real API reliability, permission handling, open-ended tool discovery, or automatic extraction of tool contracts from documentation. This controlled design enables reproducible comparison of filtering methods using task success, wrong-tool calls, premature actions, tool exposure, trajectory length, and token cost.

\section{Experimental Setup}
\label{sec:experimental_setup}

We evaluate whether causal tool filtering improves multi-step tool-use behavior under controlled conditions. All methods are run on the same task set, tool registry, mocked execution environment, and prompting protocol.

\subsection{Models and Prompting}

In the main experiment, we evaluate four tool-calling LLMs across two model families: Amazon Nova 2 Lite, Amazon Nova 2 Pro Preview, Claude 3.5 Haiku, and Claude Sonnet 4. Each model receives the user query, current symbolic state, and the visible tool schemas selected by the filtering method. The model is instructed to choose exactly one visible tool and provide valid arguments. We use deterministic decoding where supported, with a fixed maximum output length. Because the current symbolic state is supplied at every step, the controlled benchmark does not require full conversation history.

\subsection{Execution Protocol}

For each task, the agent executes a bounded tool-use loop. At step $t$, the filtering method selects a visible tool set $\mathcal{V}_t$. The model selects one tool call with tool $a_t \in \mathcal{V}_t$, the mocked environment returns a deterministic observation, and produced variables are added to the symbolic state. The loop stops when the goal state is reached, a failure occurs, or the maximum step limit is exceeded. In the main experiments, the maximum step limit is six.

A task succeeds if its goal state is reached within the step limit. A task fails if no tool is visible, the model does not call a tool, the selected tool has no valid mocked output for the task, or the step limit is reached before satisfying the goal.

\subsection{Compared Filtering Methods}

Table~\ref{tab:filtering_methods} summarizes the filtering methods.

\begin{table}[t]
\centering
\small
\begin{tabularx}{\linewidth}{l Y L{3.4cm}}
\toprule
Method & Selection rule & Purpose \\
\midrule
All tools & Expose the full registry & Tool-overload baseline \\
Keyword top-5 & Top 5 tools by keyword overlap & Relevance baseline \\
Keyword top-10 & Top 10 tools by keyword overlap & Larger relevance baseline \\
State-aware & Expose tools with $R_i \subseteq s_t$ & Executability baseline \\
Full causal path & Expose all tools on a minimal causal path & Future-tool exposure ablation \\
CMTF & Expose only the next causal frontier & Proposed method \\
\bottomrule
\end{tabularx}
\caption{Filtering methods compared in the experiments.}
\label{tab:filtering_methods}
\end{table}

The keyword baselines score tools by token overlap between the task context and tool metadata, including tool name, description, domain, required variables, and produced variables. State-aware filtering exposes every executable tool. The full-causal-path ablation exposes all tools on a minimal causal path, including future-step tools. CMTF exposes only the next executable frontier of that path.

\subsection{Evaluation Metrics}

We report the following metrics:

\begin{itemize}
    \item \textbf{Task success:} whether the goal state is reached within the step limit.
    \item \textbf{Wrong-tool count:} number of steps where the selected tool differs from the gold next tool.
    \item \textbf{Premature-action count:} number of write, send, update, share, or delete tools selected before they are appropriate for the current state.
    \item \textbf{Average tools exposed per step:} mean size of $\mathcal{V}_t$.
    \item \textbf{Trajectory length:} number of model-tool steps before success or failure.
    \item \textbf{Token cost:} total input and output tokens consumed during the task.
\end{itemize}

Each benchmark task has one intended gold chain, so wrong-tool count is measured against the gold next tool at each step. This strict metric enables controlled comparison, but may penalize alternative valid trajectories.

\subsection{Reproducibility Details}

The main experiment uses 102 tasks and a 100-tool registry. Tool execution is deterministic: each valid task-tool pair returns a predefined mocked output, and produced variables are added to the task state. Each run logs the task identifier, model, filtering method, visible tools, selected tool, gold next tool, state transition, mocked output, token usage, and error status.

\section{Results}
\label{sec:results}

We evaluate the filtering methods on a main benchmark consisting of 102 tasks, 100 available tools, four LLM backends, and six filtering strategies, yielding 2448 task-method-model runs. Table~\ref{tab:main_results} reports aggregate performance across all models and tasks.

\begin{table}[H]
\centering
\small
\begin{tabular}{lccccc}
\toprule
Method & Success & Wrong tools & Premature & Tools/step & Tokens \\
\midrule
All tools         & 0.83 & 1.25 & 0.03 & 100.00 & 24569 \\
Keyword top-5     & 0.61 & 2.36 & 0.03 & 5.00   & 4407  \\
Keyword top-10    & 0.72 & 1.93 & 0.06 & 10.00  & 5039  \\
State-aware       & 0.65 & 1.98 & 0.00 & 5.73   & 4354  \\
Full causal path  & 0.99 & 0.03 & 0.01 & 1.90   & 2555  \\
CMTF              & 0.99 & 0.01 & 0.00 & 1.00   & 2405  \\
\bottomrule
\end{tabular}
\caption{Main performance comparison across filtering methods. Success is the fraction of completed tasks. Wrong tools and premature actions are averaged per task. Tools/step denotes the average number of visible tools exposed at each decision step. Tokens denotes average token usage per task.}
\label{tab:main_results}
\end{table}

\begin{figure}[H]
\centering
\includegraphics[width=0.85\linewidth]{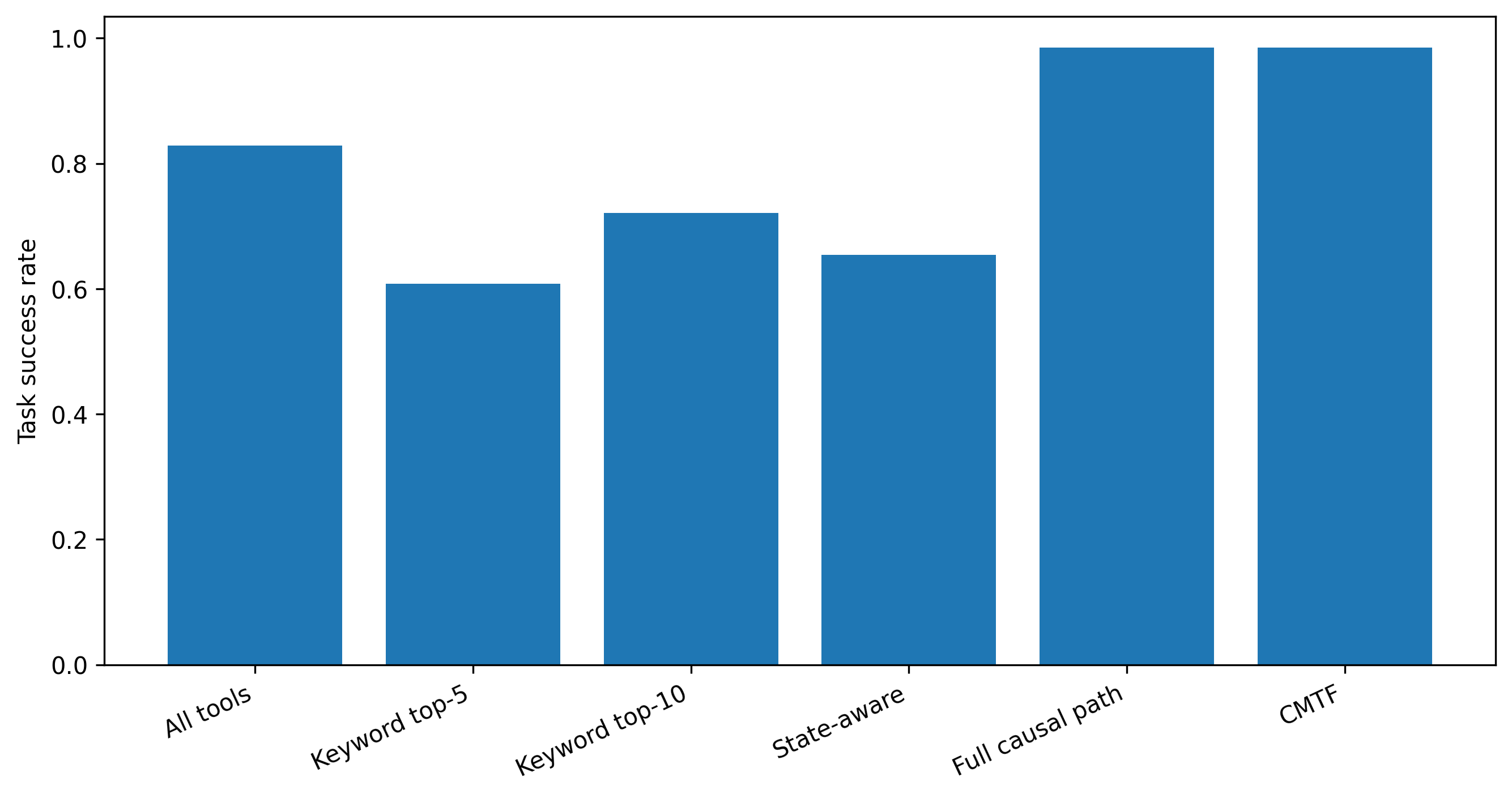}
\caption{Task success rate by filtering method. CMTF matches the strongest causal baseline while substantially outperforming semantic and state-aware filtering.}
\label{fig:success_by_method}
\end{figure}

\subsection{Aggregate Performance}

Table~\ref{tab:main_results} and Figure~\ref{fig:success_by_method} show that CMTF provides the strongest reliability-efficiency tradeoff among the evaluated methods. It achieves near-perfect aggregate success, matching the full-causal-path baseline at 0.99, while exposing only one tool per decision step. Compared with the all-tools baseline, CMTF improves success from 0.83 to 0.99, reduces wrong-tool calls from 1.25 to 0.01 per task, eliminates premature actions, and reduces average token usage from 24,569 to 2,405 tokens per task.

These results support the central claim of this paper: the relevant criterion for tool filtering is not simply semantic relevance or schema executability, but causal necessity with respect to the current task state. Exposing all tools gives the agent maximum flexibility, but also imposes a large tool-selection burden. In contrast, CMTF restricts the visible action space to the next causally necessary frontier, preserving task success while sharply reducing tool-choice errors and context cost.

\subsection{Limits of Semantic and State-Aware Filtering}

Keyword-based filtering substantially reduces the number of visible tools, but does not reliably improve task success. Keyword top-5 exposes only five tools per step, yet achieves 0.61 success and produces 2.36 wrong-tool calls per task. Keyword top-10 improves success to 0.72, but still remains below the all-tools baseline and produces 1.93 wrong-tool calls per task. This shows that semantic similarity alone can retain plausible but operationally incorrect tools or exclude tools required for causal progress.

State-aware filtering also underperforms CMTF. Although it exposes only 5.73 tools per step on average, it achieves 0.65 success and 1.98 wrong-tool calls per task. This indicates that executability is not equivalent to usefulness: a tool may have its preconditions satisfied while still being irrelevant or premature for the current step. CMTF avoids this failure mode by filtering tools according to their role in advancing the task state toward the goal.

\subsection{Causal Path Versus Causal Frontier}

The strongest baseline is full causal path exposure. It achieves the same aggregate success as CMTF, 0.99, and substantially outperforms semantic and state-aware filtering. However, full causal path exposes 1.90 tools per step on average, compared with 1.00 for CMTF. It also produces nonzero wrong-tool and premature-action rates.

This comparison isolates the benefit of exposing only the current causal frontier rather than all tools that may appear on a valid future path. Full causal path filtering confirms that causal structure is useful, while CMTF shows that the most effective interface is the minimal next-step frontier.

\subsection{Tool Exposure and Token Efficiency}

\begin{figure}[H]
\centering
\includegraphics[width=0.85\linewidth]{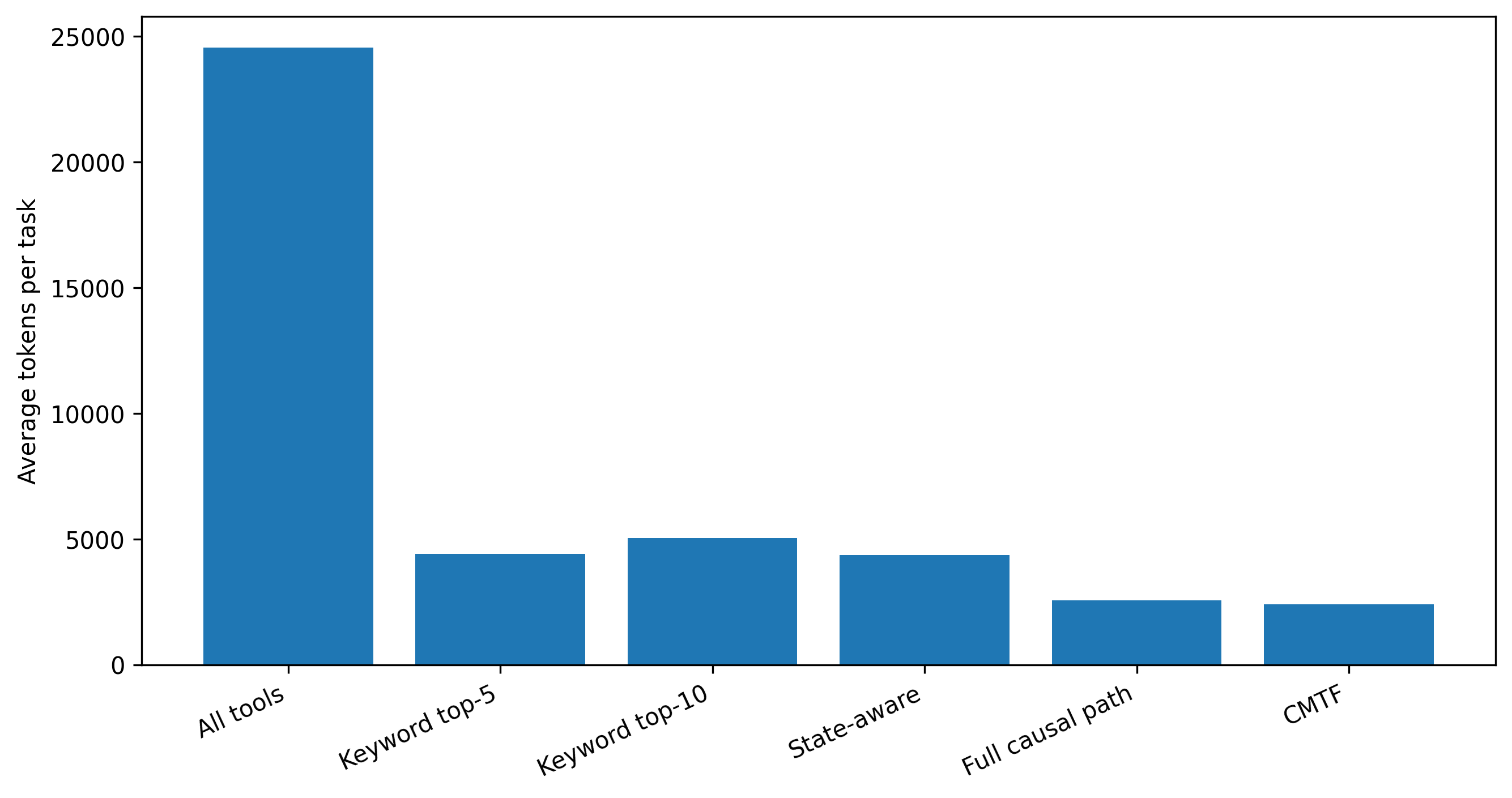}
\caption{Average token usage per task by filtering method. CMTF substantially reduces context cost relative to exposing all tools.}
\label{fig:tokens_by_method}
\end{figure}

Figure~\ref{fig:tokens_by_method} shows that CMTF substantially reduces token cost. The all-tools baseline exposes 100 tools per step and uses 24,569 tokens per task on average. CMTF exposes one tool per step and uses 2,405 tokens per task. This corresponds to roughly a 99\% reduction in visible tools and about a 90\% reduction in token usage relative to all-tools exposure.

The reduction in token cost is not obtained at the expense of reliability. CMTF has higher success and fewer wrong-tool calls than all-tools, keyword top-$k$, and state-aware filtering. Thus, causal minimal filtering improves both efficiency and behavioral reliability.

\subsection{Model-Level Effects}

The benefits of CMTF are consistent across model families. For Nova 2 Lite, all-tools and CMTF both achieve 1.00 success, but all-tools produces 1.10 wrong-tool calls and 0.11 premature actions per task, while CMTF produces none. For Nova 2 Pro Preview, CMTF improves success from 0.83 to 1.00 and eliminates wrong-tool calls.

The largest gains appear for Claude 3.5 Haiku. Under all-tools exposure, it achieves 0.48 success with 2.62 wrong-tool calls per task. With CMTF, success rises to 0.94 and wrong-tool calls fall to 0.06. Claude Sonnet 4 is strong even with all-tools exposure, reaching 1.00 success, but CMTF still removes residual wrong-tool calls and reduces token usage from 24,858 to 1,819 tokens per task.

Overall, these results show that causal minimal filtering improves weaker models by reducing tool-selection burden and improves stronger models by reducing unnecessary exposure, residual tool-choice errors, and token cost.

\section{Discussion}
\label{sec:discussion}

\subsection{Relevance Is Not Necessity}

The results highlight an important distinction between semantic relevance and causal necessity. A tool can be semantically related to a user request while still being unnecessary, redundant, or premature for the current task state. This explains why keyword top-$k$ filtering reduces the number of visible tools but does not reliably improve task success. In the main experiment, keyword top-5 and keyword top-10 reduce the visible tool set to 5 and 10 tools per step, respectively, but still underperform CMTF in both success and wrong-tool rate.

Similarly, state-aware filtering reduces the visible action space, but executability alone is insufficient. A tool may have its input preconditions satisfied while still not being the right next action. CMTF addresses this mismatch by filtering tools according to their role in advancing the current state toward the goal. This shifts tool selection from a semantic retrieval problem to a state-transition problem. The empirical results support this framing: the strongest methods are those that use causal structure, and the best reliability-efficiency tradeoff is obtained by exposing only the next causal frontier.

\subsection{Tool Exposure as a Runtime Control Surface}

Tool exposure should be treated as a runtime control surface in agentic systems. In tool-rich environments, the model is not merely reasoning over the user request; it is also reasoning over the action interface made visible to it. Exposing a large tool menu increases context length and creates additional opportunities for wrong-tool or premature-action errors.

This has direct implications for production agents. Enterprise copilots, personal assistants, and autonomous workflows often connect to many tools across email, calendars, files, databases, web search, and external APIs. In such settings, causal filtering offers a mechanism to reduce context cost, limit exposure to high-risk actions until they are justified, and make the agent's action space easier to audit. Because each exposed tool is justified by a precondition-effect relation to the goal, the filtered action set can be inspected independently of the model's internal reasoning.

The results also suggest that causal filtering can help smaller or less robust models by reducing the number of irrelevant tool decisions they must resolve. This is important for practical deployments, where cost, latency, and model availability often motivate the use of smaller models.

\subsection{Minimality--Robustness Tradeoff}

Strict minimality is most useful when the task state and tool contracts are reliable. In more uncertain settings, however, exposing only the minimal frontier may be too restrictive. If the state estimate is incomplete or a tool fails unexpectedly, the agent may need a small recovery set rather than only the nominal next-step frontier.

A practical extension is therefore to make CMTF uncertainty-aware. The default interface can expose the minimal causal frontier, while fallback or diagnostic tools can be added when execution fails, when the state tracker is uncertain, or when no causal path is found. This preserves the main benefit of minimality while allowing controlled expansion for recovery.

\subsection{Designing and Maintaining Tool Contracts}

CMTF depends on lightweight tool contracts that describe preconditions, effects, risk, and cost. In production systems, these contracts may be written manually for safety-critical tools, derived from API schemas and documentation, or inferred from execution traces. Risk annotations are especially important for tools that send, delete, update, purchase, or share information, since these actions should require stronger causal justification than read-only tools.

The experiments use structured contracts to isolate the effect of causal filtering. In practice, contract quality will affect filtering quality. This makes tool-contract design an important systems problem: better contracts should improve both reliability and auditability, while incomplete contracts may require conservative fallback policies. Overall, the results suggest that reliable tool use is not only a model capability, but also an interface-design problem between the model, the task state, and the available tools.

\section{Limitations and Threats to Validity}
\label{sec:limitations}

The benchmark is synthetic by design. This allows controlled evaluation of tool filtering under specified task states, goal states, and gold tool-chain annotations, but it may not capture the full ambiguity and variability of real tool ecosystems. In particular, the mocked tool outputs simplify environmental uncertainty such as tool failures, latency, partial results, authentication issues, and ambiguous observations. The experiments therefore primarily evaluate tool-selection behavior in simulated task environments rather than end-to-end robustness in deployed systems.

CMTF also assumes that useful tool contracts are available. Incomplete or incorrect preconditions, effects, risk annotations, or cost estimates may lead to over-filtering or incorrect tool exposure. In production settings, such contracts may need to be validated, versioned, inferred from API documentation, or monitored using execution traces.

The method further assumes that user requests can be mapped to goal states and that the current task state can be tracked with sufficient accuracy. CMTF is therefore best suited to tasks with identifiable state transitions, such as search-read-update or retrieve-summarize workflows. Open-ended, creative, exploratory, or multi-objective tasks may require more flexible goal representations and uncertainty-aware frontier expansion.

Finally, while the experiments evaluate multiple LLM backends across two model families, additional validation is needed across more providers, open-weight models, real APIs, and production tool stacks. The reported metrics focus on task success, tool-choice errors, premature actions, tool exposure, and token usage; they do not fully capture user-perceived quality, wall-clock latency, monetary cost, or the severity of safety-relevant failures.

\section{Conclusion}
\label{sec:conclusion}

This paper introduced Causal Minimal Tool Filtering (CMTF), a training-free method for reducing tool-choice confusion in tool-augmented LLM agents. Instead of exposing tools based only on semantic relevance or schema executability, CMTF uses lightweight tool contracts to identify the minimal next-step causal frontier needed to advance the current task state toward the user goal.

Across 2448 task-method-model runs with 102 tasks, 100 tools, four LLM backends, and six filtering strategies, CMTF achieved near-perfect aggregate success while substantially reducing wrong-tool calls, premature actions, visible tool exposure, and token usage. Compared with exposing all tools, CMTF reduced the visible action space from 100 tools per step to one tool per step and reduced average token usage by about 90\% while preserving near-perfect aggregate success.

The results show that simply reducing the number of tools is not sufficient: semantic top-$k$ and state-aware filtering expose smaller tool sets but still underperform causal filtering. More broadly, this work suggests that reliable tool-augmented agents require tool interfaces that are dynamically shaped by task state, causal structure, and risk, rather than tool menus optimized only for relevance.

\section*{Acknowledgments}
The authors thank colleagues for helpful feedback. This work was conducted in the authors' personal capacity. The views expressed in this paper are solely those of the authors and do not necessarily reflect the views of their employers.
\section*{Funding}
This work did not receive external funding.
\section*{Conflicts of Interest}
The authors declare no conflicts of interest.
\section*{Artifact Availability}

The synthetic benchmark, tool registry, filtering implementations, evaluation scripts, and analysis utilities are available at:

\begin{center}
\url{https://github.com/R-Suresh/ToolChoiceConfusion}
\end{center}.
\bibliographystyle{plain}
\bibliography{references}
\end{document}